\documentclass[10pt,twocolumn,letterpaper]{article}

\usepackage[pagenumbers]{cvpr} 

\usepackage{graphicx}
\usepackage{amsmath}
\usepackage{amssymb}
\usepackage{booktabs}
\usepackage{multirow}
\usepackage{makecell}
\usepackage{floatrow}
\usepackage{caption}
\usepackage{xcolor}
\makeatletter
\@namedef{ver@everyshi.sty}{}
\makeatother
\usepackage{tikz}
\usepackage{enumitem}
\usepackage{tikzpagenodes}
\definecolor{citecolor}{HTML}{2980b9}
\definecolor{linkcolor}{HTML}{c0392b}
\usepackage{stfloats}

\usepackage{xcolor}
\usepackage{color, colortbl}
\definecolor{zrrgreen}{HTML}{008000}
\definecolor{zrrblue}{HTML}{4682B4}
\definecolor{zrrred}{HTML}{B22222}

\usepackage{adjustbox}
\usepackage{colortbl}

\makeatletter
  \newcommand\figcaption{\def\@captype{figure}\caption}
  \newcommand\tabcaption{\def\@captype{table}\caption}
\makeatother
\usepackage[hidelinks,pagebackref=true,breaklinks=true,colorlinks,bookmarks=false,citecolor=citecolor,letterpaper=true,linkcolor=linkcolor]{hyperref}

\usepackage[capitalize]{cleveref}
\crefname{section}{Sec.}{Secs.}
\Crefname{section}{Section}{Sections}
\Crefname{table}{Table}{Tables}
\crefname{table}{Tab.}{Tabs.}
\newcommand\blfootnote[1]{%
  \begingroup
  \renewcommand\thefootnote{}\footnote{#1}%
  \addtocounter{footnote}{-1}%
  \endgroup
}

\usepackage{framed}
\usepackage{makecell}
\usepackage{listings}
\definecolor{lightgray}{rgb}{.9,.9,.9}
\definecolor{darkgray}{rgb}{.4,.4,.4}
\definecolor{purple}{rgb}{0.65, 0.12, 0.82}
\lstdefinelanguage{JavaScript}{
  keywords={break, case, catch, continue, debugger, default, delete, do, else, false, finally, for, function, if, in, instanceof, new, null, return, switch, this, throw, true, try, typeof, var, void, while, with},
  morecomment=[l]{//},
  morecomment=[s]{/*}{*/},
  morestring=[b]',
  morestring=[b]",
  ndkeywords={class, export, boolean, throw, implements, import, this},
  keywordstyle=\color{blue}\bfseries,
  ndkeywordstyle=\color{darkgray}\bfseries,
  identifierstyle=\color{black},
  commentstyle=\color{purple}\ttfamily,
  stringstyle=\color{red}\ttfamily,
  sensitive=true
}
\def\cvprPaperID{XXXX} 
\def\confName{CVPR}
\def\confYear{2023}

\begin{document}

\title{LLaMA-Adapter V2: Parameter-Efficient Visual Instruction Model}

\author{
Peng Gao$^{*\ddagger\dag1}$, Jiaming Han$^{*1}$, Renrui Zhang$^{*1,2}$, Ziyi Lin$^{*2}$, Shijie Geng$^{3}$, Aojun Zhou$^{2}$ \\ Wei Zhang$^{1}$, Pan Lu, Conghui He$^{1}$, Xiangyu Yue$^{2}$, Hongsheng Li$^{\dag2}$, Yu Qiao$^{\dag1}$\vspace{0.2cm}\\
$^1$Shanghai Artificial Intelligence Laboratory\quad 
$^2$CUHK MMLab\\
$^3$Rutgers University\\
\texttt{\small \{gaopeng, hanjiaming, zhangrenrui, qiaoyu\}@pjlab.org.cn}
}
\maketitle

\blfootnote{$^\ddagger$Project leader, $^*$ Equal contribution, $^\dag$ Corresponding author}

\begin{tikzpicture}[remember picture,overlay,shift={(current page.north west)}]
\node[anchor=north west,xshift=1.4cm,yshift=-2.8cm]{\includegraphics[width=1.5cm]{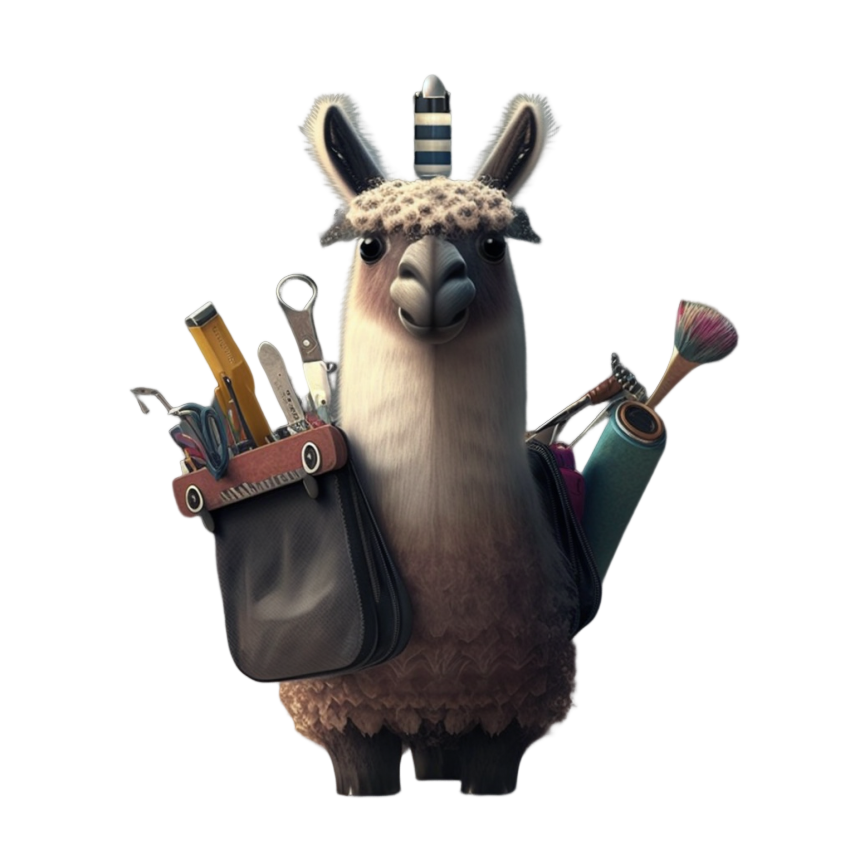}};
\end{tikzpicture}

\vspace{-0.2cm}
\begin{abstract}
\vspace{-0.2cm}
How to efficiently transform large language models (LLMs) into instruction followers is recently a popular research direction, while training LLM for multi-modal reasoning remains less explored.~Although the recent LLaMA-Adapter demonstrates the potential to handle visual inputs with LLMs, it still cannot generalize well to open-ended visual instructions and lags behind GPT-4.~In this paper, we present LLaMA-Adapter V2, a parameter-efficient visual instruction model. Specifically, we first augment LLaMA-Adapter by unlocking more learnable parameters (\emph{e.g.}, norm, bias and scale), which distribute the instruction-following ability across the entire LLaMA model besides adapters. Secondly, we propose an early fusion strategy to feed visual tokens only into the early LLM layers, contributing to better visual knowledge incorporation.
Thirdly, a joint training paradigm of image-text pairs and instruction-following data is introduced by optimizing disjoint groups of learnable parameters. This strategy effectively alleviates the interference between the two tasks of image-text alignment and instruction following and achieves strong multi-modal reasoning with only a small-scale image-text and instruction dataset. During inference, we incorporate additional expert models (\emph{e.g.} captioning/OCR systems) into LLaMA-Adapter to further enhance its image understanding capability without incurring training costs. Compared to the original LLaMA-Adapter, our LLaMA-Adapter V2 can perform open-ended multi-modal instructions by merely introducing 14M parameters over LLaMA. The newly designed framework also exhibits stronger language-only instruction-following capabilities and even excels in chat interactions. Our code and models are available at~\url{https://github.com/ZrrSkywalker/LLaMA-Adapter}.

\end{abstract}

\begin{figure}[t]
  \centering
\includegraphics[width=\textwidth]{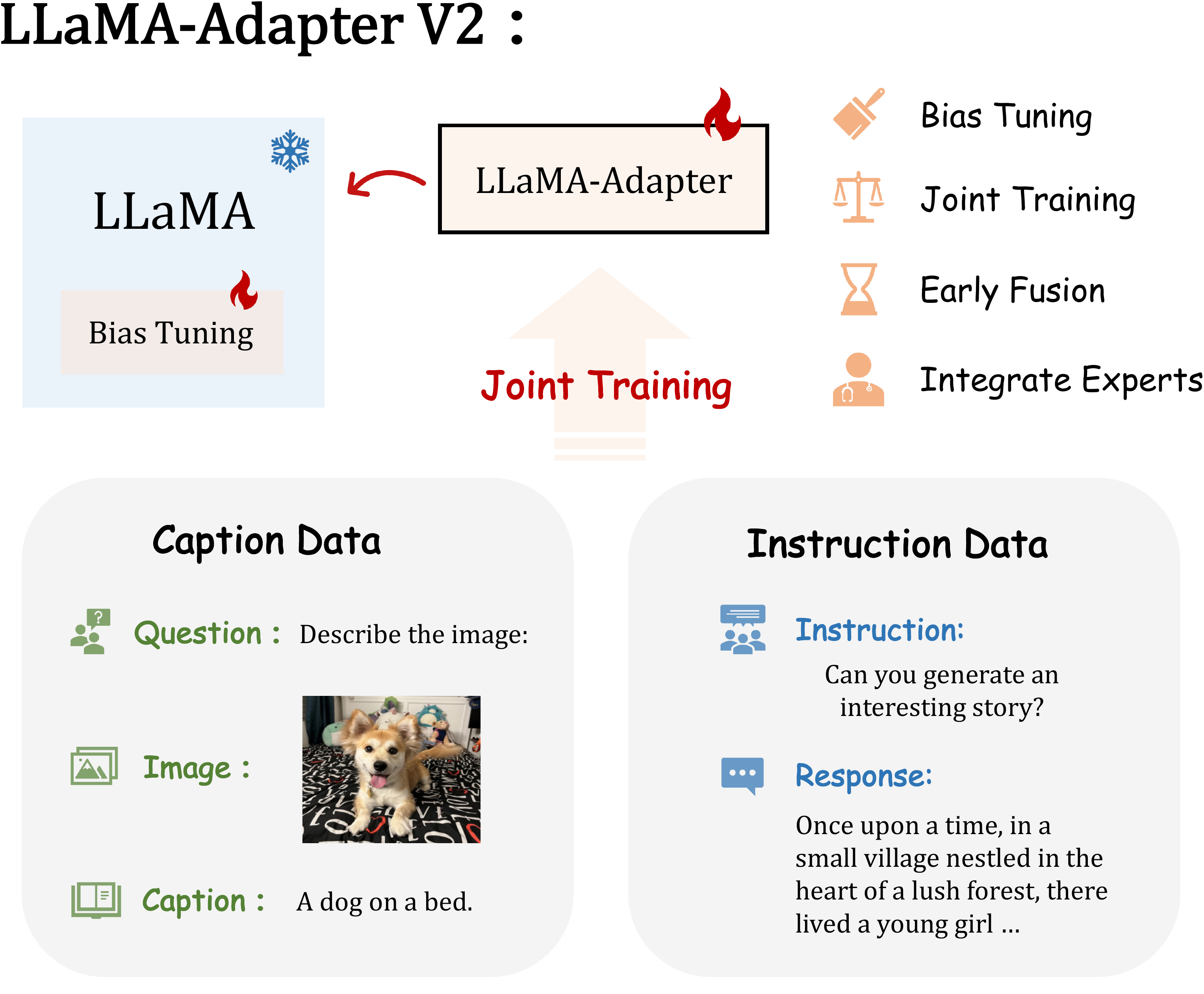}
   \caption{\textbf{Training Pipeline of LLaMA-Adapter V2.} We introduce several strategies to enhance the capability of LLaMA-Adapter~\cite{llamaadapter2023}, which enable a parameter-efficient visual instruction model with superior multi-modal reasoning.}
    \label{fig1}
\vspace{0.2cm}
\end{figure}

\section{Introduction}
\label{sec:intro}

Large Language Models (LLMs) \cite{zhao2023survey} have garnered significant attention in the AI community because of their exceptional ability to comprehend, reason with, and generate human language.~To make LLM's responses more vivid and convincing, recent works \cite{alpaca,vicuna2023,peng2023instruction,koala_blogpost_2023,xu2023baize} have explored transforming LLMs into instruction-following models.~For example, Stanford Alpaca \cite{alpaca} fine-tunes LLaMA \cite{touvron2023llama} into an instruction-following model using instruction examples generated from OpenAI's InstructGPT model \cite{ouyang2022training}. Follow-up works of Alpaca further extend LLaMA by utilizing higher-quality instruction data, such as ShareGPT~\cite{sharegpt} and those generated by GPT-4~\cite{peng2023instruction}. Compared with the full fine-tuning paradigm adopted by Alpaca and Vicuna~\cite{vicuna2023}, LLaMA-Adapter~\cite{llamaadapter2023} introduces lightweight adapters with zero-initialized attention into the frozen LLaMA for parameter-efficient fine-tuning, along with multi-modality knowledge injection. Despite the significant progress, the aforementioned approaches are still unable to perform more advanced multi-modal instructions, \emph{e.g.}, visual understanding like GPT-4 \cite{OpenAI2023GPT4TR}.

Most recently, studies such as MiniGPT-4 \cite{zou2022xdecoder} and LLaVA \cite{liu2023visual} have sparked a new wave of research on extending language-only instruction models into multi-modal ones to empower LLMs with visual reasoning ability, in a similar way to LLaMA-Adapter. MiniGPT-4 connects a frozen visual encoder and an LLM by pre-training on 134 million image-text pairs, and then improves the model's performance by further fine-tuning on a well-aligned image-text dataset. LLaVA also leverages image-text pairs to align the visual model and the LLM. Different from MiniGPT-4, LLaVA fine-tunes the whole LLM on 150K high-quality multi-modal instruction data generated by GPT-4. While these methods demonstrate impressive multi-modal understanding capabilities, they require updating billions of model parameters and meticulously collecting a substantial amount of multi-modal training data, which are either annotated by humans or distilled from responses of OpenAI API.

In this paper, we aim to design a parameter-efficient visual instruction model. We build on the popular parameter-efficient LLaMA-Adapter to develop our new method, which we refer to as LLaMA-Adapter V2.~LLaMA-Adapter was initially developed as an instruction-following model and can be easily transformed into a visual instruction model by incorporating visual features into the adaptation prompts. However, due to the lack of multi-modal instruction tuning data, the multi-modal version of LLaMA-Adapter is restricted as a traditional vision-language model. For instance, LLaMA-Adapter trained on COCO Caption \cite{chen2015microsoft} can only produce short image captions when given a specific prompt, such as ``Generate caption for this image". The model cannot adapt to open-ended multi-modal instructions, such as complex visual reasoning and visual question answering tasks.


Although we do not currently leverage multi-modal instruction data, it is still possible to perform multi-modal instruction tuning for LLaMA-Adapter. We begin by using a frozen instruction-following LLaMA-Adapter model as the starting point and refine it by optimizing the visual projection layers on image-text pairs to ensure proper vision-language alignment. However, we have observed that the visual features tend to dominate the adaptation prompts, causing the instruction-following ability to deteriorate rapidly.

To address this challenge, we propose a simple \textbf{\textit{early fusion of visual knowledge}} strategy that resolves the interference between the two tasks of image-text alignment and language instruction tuning. In LLaMA-Adapter, dynamic visual prompts are incorporated into the static adaptation prompts at the last $L$ layers.~However, in LLaMA-Adapter V2, we distribute the dynamic visual prompts to only the first $K$ layers, where $K<N-L$ and $N$ denote the total number of Transformer layers.~As a result, image-text alignment no longer disrupts the model's instruction-following capability. 
With this strategy, we can achieve superior visual instruction learning through the \textbf{\textit{joint training with disjoint parameters}} with image captioning data and instruction-following data, even in the absence of high-quality multi-modal instruction data.~Additionally, we augment LLaMA-Adapter by unlocking more learnable parameters such as normalization, layer bias and scale, which we refer to as \textbf{\textit{bias tuning of linear layers}}.~By increasing the model's tunable capacity, we can spread the instruction-following knowledge across the entire LLM. It is worth noting that these parameters only account for approximately 0.04\% of the entire model, ensuring that LLaMA-Adapter V2 remains a parameter-efficient approach.


Finally, we introduce additional expert models (\emph{e.g.}, captioning, detection, and OCR systems) to enhance the image understanding capabilities of LLaMA-Adapter V2, setting our approach apart from others such as MiniGPT-4 and LLaVA that rely on massive image-text pair training data. By collaborating with these specialized experts, our framework gains increased flexibility and allows plugging in various experts for a wide variety of tasks without the necessity of pre-training on extensive vision-language data.

\begin{table*}[t]
\centering
\small
\resizebox{\textwidth}{!}{%
\begin{tabular}{c|ccc|cc|cc|c}
\toprule
\multirow{2}{*}{Model} & \multicolumn{3}{c}{Language Instruction Data} & \multicolumn{2}{c}{Image-Text Data} & \multicolumn{2}{c}{Visual Instruction Data} & Tuning Parameters \\
                       & Source        & Type              & Size      & Source                   & Size     & Source                  & Size              & Size            \\
\midrule
MiniGPT-4~\cite{zou2022xdecoder}               & ShareGPT~\cite{sharegpt}      & Conversation      & 70K       & CC-VG-SBU-L400$^*$      & 134M       & CC+ChatGPT              & 5K                & 13B     \\
LLaVA~\cite{liu2023visual}                  & ShareGPT~\cite{sharegpt}     & Conversation      & 70K       & CC$^*$                   & 595K     & COCO+GPT4               & 158K              & 13B     \\
\textbf{LLaMA-Adapter V2}       & GPT-4-LLM~\cite{peng2023instruction}      & Single-turn       & \textbf{52K}       & COCO~\cite{chen2015microsoft}                     & \textbf{567K}     & -                       & \textbf{0}                 & \textbf{14M}\\
\bottomrule
\end{tabular}%
}
\caption{\textbf{Training Comparison of Different Methods.} CC, VG and L400 represent Conceptual Caption~\cite{sharma2018conceptual,changpinyo2021conceptual}, Visual Genome~\cite{krishna2017visual} and LAION 400M~\cite{schuhmann2021laion}, respectively. Note that we count all the data and tuning parameters needed to convert the pretrained vision model and LLM into a visual instruction model. $*$ denotes the filtered dataset.}
\label{tab:param_compare_with_minigpt4_llava}
\end{table*}

Fig. \ref{fig1} and Fig. \ref{fig2} illustrate the whole training and generation pipelines of our LLaMA-Adapter V2, respectively. We summarize our main contributions as follows:
\begin{itemize}
    \item \textbf{Stronger Language Instruction Model.}~With the parameter-efficient tuning strategy and high-quality language instruction data, LLaMA-Adapter V2 surpasses its predecessor LLaMA-Adapter in terms of language instruction-following performance. Moreover, LLaMA-Adapter V2 is capable of conducting multi-turn dialog, demonstrating its stronger ability as a language instruction model.
    
    \item \textbf{Balanced Visual Instruction Tuning.} We propose a simple early fusion strategy to solve the interference between image-text alignment and instruction-following learning targets. As a result, we transform LLaMA-Adapter V2 into a visual instruction model without the need of multi-modal instruction training data.
    
    \item \textbf{Integration of Expert Systems.} Instead of end-to-end pre-training on massive image-text pairs, we embrace modular design where different expert models can be integrated into our framework to enhance the image understanding abilities of LLMs.
\end{itemize}

\section{Related Work}
\label{sec:related}

\paragraph{Instruction-following Language Models}
Large Language Models (LLMs) are pre-trained on extensive text corpora using autoregressive Transformer models to predict subsequent tokens.~They have demonstrated strong capabilities as self-supervised \cite{radford2018improving}, multi-task \cite{radford2019language}, and few-shot learners \cite{brown2020language}.~Recently, InstructGPT \cite{ouyang2022training} and FLAN \cite{wei2022finetuned,chung2022scaling} have shown that LLMs can be converted into instruction-following models by fine-tuning them on instructional datasets.~To facilitate the generation of instruction-following examples, Self-Instruct \cite{wang2022self} employs a semi-automated, iterative bootstrapping algorithm that expands a limited seed set of manually-written instructions and progressively expanding the collection of tasks using off-the-shelf LLMs.~Alpaca \cite{alpaca} applies the Self-Instruct strategy to produce 52K high-quality instruction-following demonstrations and fine-tunes upon the open-source LLaMA \cite{touvron2023llama} model, ultimately obtaining an instruction-following language model exhibiting many behaviors similar to OpenAI's text-davinci-003. Inspired by Alpaca's success, Vicuna \cite{vicuna2023} and GPT-4-LLM \cite{peng2023instruction} further reveal that dialog and enhanced instruction-following capabilities can be ignited by fine-tuning on either user-shared ChatGPT conversations or instruction-following data generated by the GPT-4 API. 
However, Alpaca, Vicuna and GPT-4-LLM all fine-tune the full parameters of LLMs, resulting in unaffordable GPU memory usage and training costs. In contrast, LoRA \cite{hu2022lora} and LLaMA-Adapter \cite{llamaadapter2023} validate that parameter-efficient fine-tuning approaches can potentially replace full parameter updates during supervised fine-tuning of LLMs. In this paper, LLaMA-Adapter V2 goes a step further by constructing a parameter-efficient zero-shot visual instruction model that reuses the instruction-following ability of LLaMA-Adapter.

\paragraph{Visual Instruction Models}
Traditional image captioning~\cite{vinyals2016show,lu2017knowing, anderson2018bottom, hossain2019comprehensive, cornia2020meshed,li2023blip} and visual question answering (VQA)~\cite{gao2019dynamic, lu2019vilbert, li2019visualbert, pmlr-v139-kim21k, scienceqa, Min2021FILMFI, zhou2020unified, goyal2017making, alayrac2022flamingo} approaches can be viewed as simplified versions of visual instruction models. However, their image understanding capabilities fall short compared to GPT-4. 
To be specific, image captioning models can only describe images using concise, short sentences, while VQA systems are capable of answering various visual-related questions but without providing explicit explanations. Therefore, these systems have yet to achieve the level of proficiency required to meet the demands of real-world applications. Recently, GPT-4 has showcased remarkable visual instruction-following abilities by handling mixed inputs of images and text for diverse tasks, ranging from meme explanation, document-level reasoning, exam problem-solving, and so on. In pursuit of developing visual instruction-following abilities akin to GPT-4, both LLaVA \cite{liu2023visual} and MiniGPT-4 \cite{zou2022xdecoder} assemble high-quality multi-modal instruction-following data using ChatGPT or GPT-4. They then integrate visual encoders and LLMs via a projection layer and fine-tune the models on the curated data. Diverging from these approaches, LLaMA-Adapter V2 strives to build a parameter-efficient visual instruction model without relying on multi-modal instruction data. Rather, LLaMA-Adapter V2 can function effectively using just language instruction data and image-text pairs.

\paragraph{Parameter-efficient Fine-tuning~} The pre-training-fine-tuning paradigm has been proven highly effective in various tasks such as visual recognition, language understanding, text generation, and image synthesis from textual descriptions. However, as model sizes continue to increase exponentially, fine-tuning every parameter in a colossal model becomes increasingly impractical.~In contrast, parameter-efficient fine-tuning (PEFT) approaches \cite{peft,ding2022delta} freeze most of the parameters of foundation models and only optimize a small portion of them.~Many successful PEFT approaches \cite{sung2022vl,he2022towards,sung2022lst,gao2021making,jia2022visual,lester2021power,stable-diffusion-lora,zhang2022pointclip,zhang2023prompt} have been proposed for adopting popular pre-trained models such as BERT \cite{devlin2019bert}, GPT \cite{radford2018improving,radford2019language,brown2020language}, ViT \cite{dosovitskiy2021an}, CLIP \cite{radford2021learning}, and Stable Diffusion \cite{rombach2022high} to various downstream tasks. In general, these PEFT approaches can be divided into three categories, namely prefix tuning (\emph{e.g.} \cite{li2021prefix,zhou2022coop}), reparameterization (\emph{e.g.} \cite{hu2022lora,mahabadi2021compacter,stable-diffusion-lora}), and adapters (\emph{e.g.} \cite{houlsby2019parameter,gao2021clip,zhang2022tip}).
In this paper, we present LLaMA-Adapter V2, an elegant and efficient marriage of prefix-tuning and adapter techniques. By utilizing an early fusion strategy and bias tuning, LLaMA-Adapter V2 injects visual features into large language models, yielding impressive multi-modal instruction-following performance with only 0.04\% parameters of the entire LLaMA. 


\paragraph{Integration of Expert Systems}
Collaboration among multiple expert models has proven to be a successful practice in the field of AI, as it often leads to improved performance and robustness. This is especially true in the realm of computer vision tasks, where ensembles of multiple models consistently achieve top positions on the challenge leaderboards. 
In recent years, with the continual expansion of LLMs' capabilities, there has been a growing trend \cite{qin2023tool,wu2023visual,shen2023hugginggpt,ge2023openagi,lu2023chameleon,yang2023mm,lin2023text2motion,driess2023palm,huang2022inner,gupta2022visual,suris2023vipergpt,zhu2022pointclip} to combine them with visual foundation models and leverage their combined strengths to tackle more complex vision-language tasks. By utilizing LLMs as a core controller for external visual models, these experts in turn support LLMs to perform a wider range of tasks that demand a deeper visual understanding. For example, recent studies such as HuggingGPT \cite{shen2023hugginggpt}, Visual ChatGPT \cite{wu2023visual}, Chameleon \cite{lu2023chameleon}, MMReACT \cite{yang2023mm}, and  ViperGPT \cite{suris2023vipergpt} utilize LLMs as the central manager to perform compositional task planning and call upon off-the-shelf expert models/tools to aid in various complex multi-modal tasks, including understanding, generation, search, reasoning, and programming, etc.~In addition, PaLM-E \cite{driess2023palm}, Inner Monologue \cite{huang2022inner}, and Text2Motion \cite{lin2023text2motion} further extend the capabilities of LLMs to robotics by incorporating real-world sensor modalities. These embodied LLMs possess the power to comprehend natural language instructions and execute sequential manipulation planning in the real world.~As a result, the aforementioned approaches facilitate the seamless integration of different expert systems and boost the overall performance and capabilities of LLMs.
LLaMA-Adapter V2 stands out by integrating short, yet precise descriptions generated by LLaMA-Adapter during inference time for efficient zero-shot and training-free visual instruction understanding, setting it apart from other methods that require extensive multi-modal data during training. In the future, more expert visual systems will be integrated into LLaMA-Adapter V2 for building stronger visual instruction models.

\section{A Revisit of LLaMA-Adapter}
\label{sec:preliminary}


\paragraph{Zero-initialized Attention.}
As a parameter-efficient fine-tuning solution for adapting LLaMA to acquire instruction-following capability, LLaMA-Adapter~\cite{llamaadapter2023} freezes the entire LLaMA model~\cite{touvron2023llama} and introduces only an extra lightweight adapter module with 1.2M parameters.~The adapter layers are employed at the higher Transformer layers of LLaMA and concatenate a set of learnable soft prompts as the prefix to the word tokens. To incorporate newly adapted knowledge into the frozen LLaMA, LLaMA-Adapter proposes a zero-initialized attention mechanism, which enables adaptively controlling the contribution of adaptation prompts to the word tokens by learning a gating factor initialized by zero.
The gating magnitude progressively increases during training, thereby gradually injecting the instruction-following ability into the frozen LLaMA.
This strategy not only preserves LLaMA's language generation ability during the early training stages but also continuously incorporates new knowledge to enable a powerful instruction follower.


\paragraph{Simple Multi-modal Variant.}
Besides fine-tuning using language-only instructions, LLaMA-Adapter can also incorporate image and video inputs for multi-modal reasoning. For instance, when dealing with images, LLaMA-Adapter employs a pre-trained visual encoder such as CLIP~\cite{radford2021learning} to extract multi-scale visual features.
These features are then aggregated into a global feature and passed through a learnable projection layer to align the visual semantics with linguistic embedding space. Afterward, the global visual feature is added element-wisely to every adaptation prompt at the higher layers of the Transformer. 
This allows LLaMA-Adapter to generate responses based on both textual and visual inputs, resulting in competitive performance on the ScienceQA benchmark~\cite{scienceqa}.

\paragraph{Open-ended Multi-modal Reasoning.}
While LLaMA-Adapter is capable of handling relatively simple tasks such as ScienceQA, it is still unclear whether it can generate open-ended responses, such as those required for general-purpose visual question answering.
To investigate this, we first start with a LLaMA-Adapter pre-trained on language instruction data for leveraging its existing instruction-following capabilities. 
We then conduct experiments by fine-tuning its adapter modules and visual projection layers on the COCO Caption~\cite{chen2015microsoft} dataset. 
However, we found that the newly learned visual cues tend to dominate the adaptation prompts, overriding the inherent instruction-following characteristics.~Therefore, we propose LLaMA-Adapter V2, a parameter-efficient visual instruction model, to fully unleash the multi-modal potential of LLaMA.

\section{LLaMA-Adapter V2}
\label{sec:llama_adapter_v2}

%
In this section, we present the technical details of LLaMA-Adapter V2, including \textit{\textbf{bias tuning of linear layers}} (Sec.~\ref{subsec:bias_tuning}) to enhance its language instruction-following ability, \textit{\textbf{joint training with disjoint parameters}} (Sec.~\ref{subsec:joint}) for balanced visual instruction tuning, \textit{\textbf{early fusion of visual knowledge}} (Sec.~\ref{subsec:early_fusion}) to balance textual and visual understanding, and \textit{\textbf{integration with experts}} (Sec.~\ref{subsec:with_experts}) to boost zero-shot multi-modal reasoning.

\subsection{Bias Tuning of Linear Layers}
\label{subsec:bias_tuning}

LLaMA-Adapter employs learnable adaptation prompts with the zero-initialized attention mechanism (Sec.~\ref{sec:preliminary}) on the frozen LLaMA model, which allows for efficient incorporation of new knowledge. However, the parameter updates are limited to the adaptation prompts and the gating factor, without modifying the internal parameters of LLMs, which restricts its ability to perform deep fine-tuning.
In light of this, we propose a bias tuning strategy to further fuse instruction cues into LLaMA besides the adaptation prompts and the gating factor.
Specifically, to adaptively handle the tasks of instruction-following data, we first unfreeze all the normalization layers in LLaMA. For each linear layer in the Transformer, we add a bias and a scale factor as two learnable parameters. We denote the input and pre-trained weights of a certain linear layer as $\mathbf{x}$ and $\mathbf{W}$, respectively. In LLaMA-Adapter V2, we modify the linear layer using the bias $b$ and scale $s$ as
\begin{align}
    \mathbf{y} = \mathbf{W\cdot x}\ \ \ \rightarrow\ \ \ \mathbf{y} = s\cdot (\mathbf{W\cdot x} + b),\\
    \text{where}\ \ \ b = \operatorname{Init}(0), \ \ s = \operatorname{Init}(1).
\end{align}
Similar to zero-initialized attention, we initialize the bias and scale factors with zeros and ones, respectively, to stabilize the training process at early stages.
With the incorporation of the bias tuning strategy and high-quality instruction data~\cite{peng2023instruction}, LLaMA-Adapter V2 acquires superior instruction-following capabilities. Notably, the number of newly added parameters only accounts for 0.04\% ($\sim$5M) of the entire LLaMA, showing that LLaMA-Adapter V2 is still a highly parameter-efficient approach. 

\begin{figure*}[t!]
    \centering
    \includegraphics[width=\textwidth]{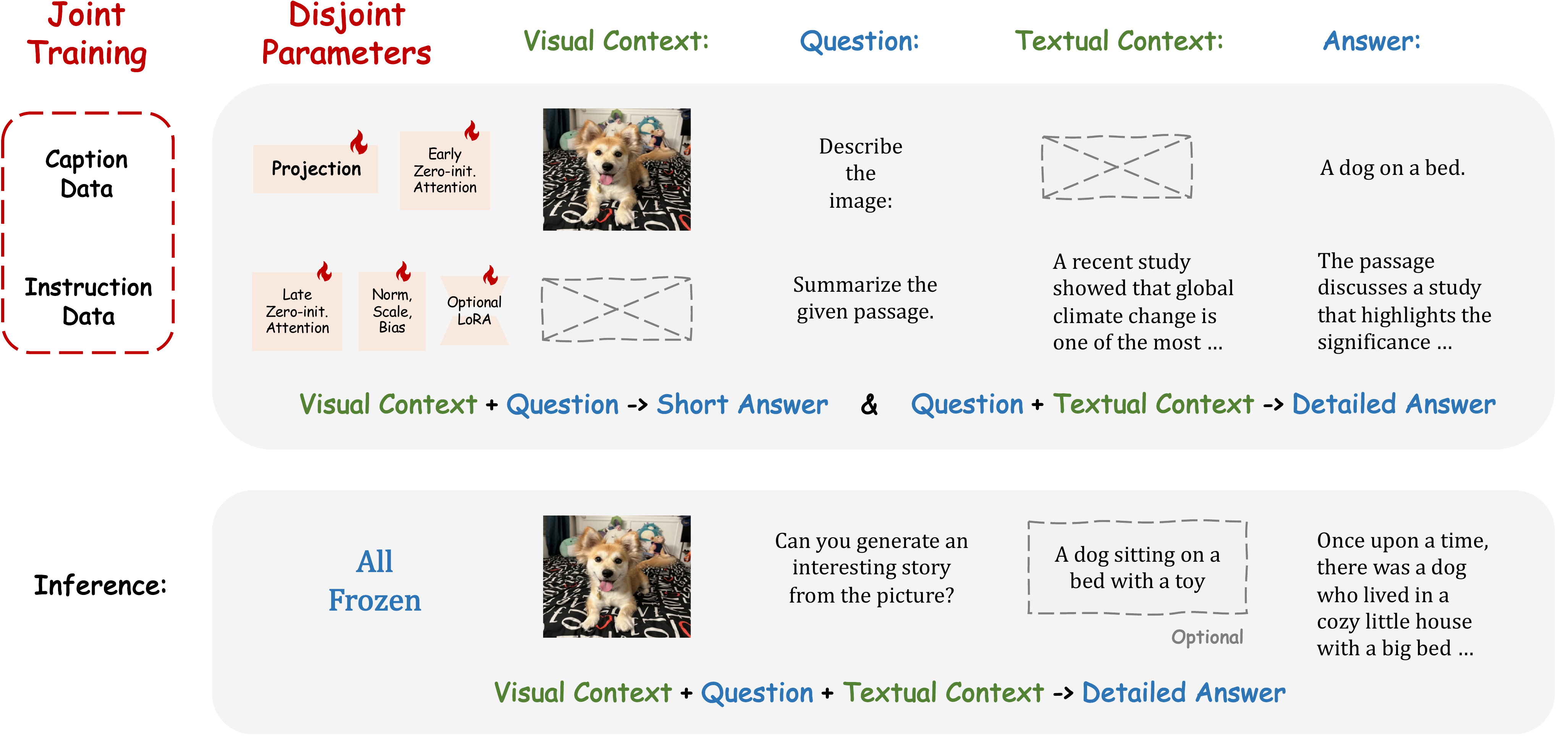}
    \caption{\textbf{Joint Training Paradigm in LLaMA-Adapter V2.} We utilize both image-text caption and language-only instruction data to jointly train LLaMA-Adapter V2, optimizing disjoint groups of learnable parameters.}
    \label{fig22}
\end{figure*}

\paragraph{Discussion.} Our bias tuning strategy bears similarity to prior parameter-efficient methodologies such as BitFit \cite{zaken2021bitfit} for BERT fine-tuning and SSF \cite{lian2022scaling} for visual prompt tuning \cite{jia2022visual}. 
However, both BitFit and SSF are designed for comprehension tasks with an 80-million parameter scale, whereas our bias tuning demonstrates its efficiency on large language models ranging from 7 billion to 65 billion parameters, such as LLaMA and GPT-3.
Moreover, our bias tuning strategy is input-agnostic, unlike Low-Rank Adaptation (LoRA) that adds an input-aware bias using low-rank transformation, further reducing the fine-tuning cost.


\subsection{Joint Training with Disjoint Parameters}
\label{subsec:joint}
Our goal is to simultaneously endow LLaMA-Adapter V2 with the capabilities of generating long language responses and multi-modal understanding. As shown in Fig.~\ref{fig22}, we propose a joint training paradigm for LLaMA-Adapter V2 to leverage both image-text captioning data and language-only instruction examples.
Due to the data volume difference between 500K image-text pairs and 50K instruction data, naively combining them for optimization can severely harm LLaMA-Adapter's instruction-following ability, just as discussed in Sec.~\ref{sec:preliminary}. 
Therefore, our joint training strategy optimizes disjoint groups of parameters in LLaMA-Adapter V2 for image-text alignment and instruction-following respectively. 
Specifically, only the visual projection layers and early zero-initialized attention with gating are trained for image-text captioning data, while the late adaptation prompts together with zero gating, the unfrozen norm, newly added bias and scale factors (or optional low-rank adaption~\cite{hu2021lora}) are utilized for learning from the instruction-following data. Such disjoint parameter optimization naturally solves the interference issue between image-text understanding and instruction following, which contributes to the emergent visual instruction-following ability of LLaMA-Adapter V2.

\paragraph{Discussion.}
Aided by our joint training strategy, LLaMA-Adapter V2 requires no high-quality multi-modal instruction data like MiniGPT-4~\cite{zou2022xdecoder} and LLaVA~\cite{liu2023visual}, but only image-text pairs and instruction-following data, as compared in Tab.~\ref{tab:param_compare_with_minigpt4_llava}. The captioning data extends LLMs for image understanding with short answers as shown in Fig.~\ref{fig22}, 
while the language-only instruction data is utilized to preserve LLaMA's capacity to generate long detailed sentences. With such complementarity, LLaMA-Adapter V2 achieves superior multi-modal reasoning by solely small-scale image-text and instruction-following data, without the need of high-quality multi-modal instruction data.

\begin{figure}
    \centering
    \vspace{0.1cm}
    \includegraphics[width=\textwidth]{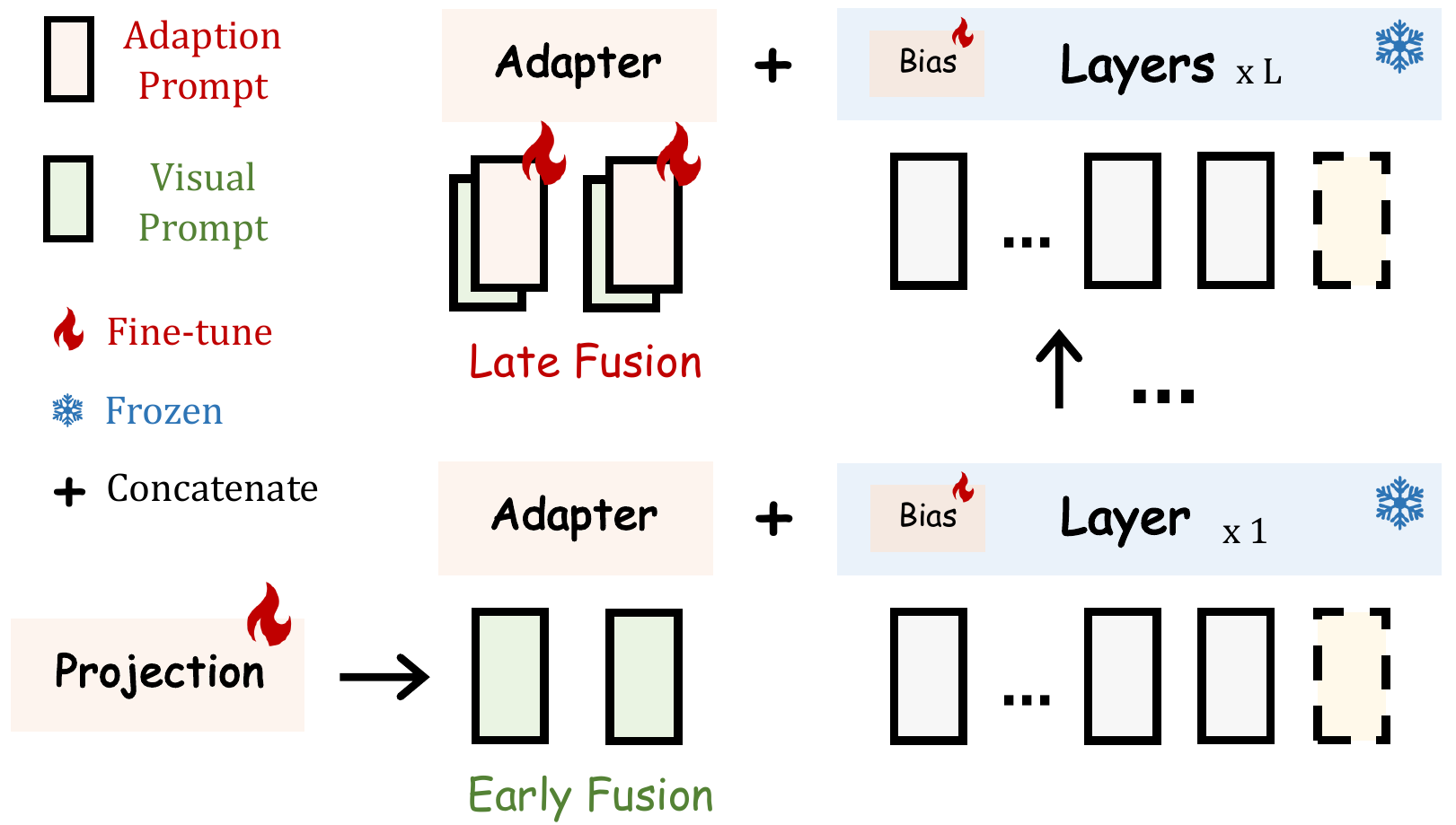}
    \caption{\textbf{Early Fusion of Visual Knowledge.} Following LLaMA-Adapter, we insert static adaptation prompts into the last $L$ layers. For visual prompts, we insert them in the early stage of LLM, disjointing with adaptation prompts.}
    \label{fig:early_fusion}
    \vspace{0.2cm}
\end{figure}

\begin{figure}[t]
  \centering
\includegraphics[width=0.9\textwidth]{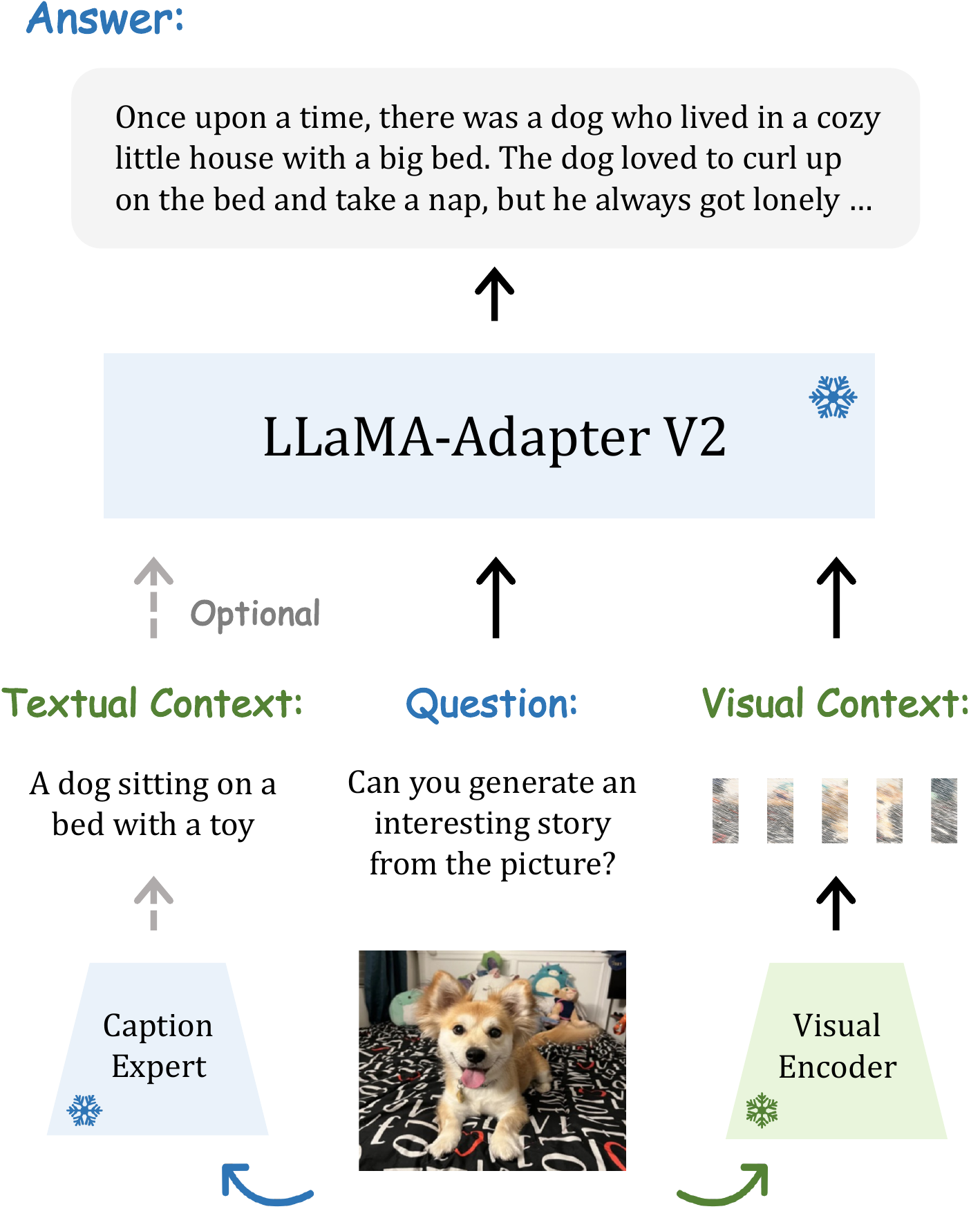}
   \caption{\textbf{Generation Pipeline of LLaMA-Adapter V2.} During inference, we introduce additional caption experts to generate a textual context for the input image, demonstrating strong visual understanding capacity of LLaMA-Adapter V2.}
    \label{fig2}
\vspace{0.2cm}
\end{figure}

\begin{table*}[th]
\centering
\small
\footnotesize
\begin{tabular}{p{1.0\columnwidth}}

\begin{framed}
\centering
 \begin{minipage}{\textwidth}
{\bf Instruction:} How can I develop my critical thinking skills? Give me three suggestions. \\
{\bf LLaMA-Adapter V2:} \\
1. Read widely: Reading books, articles, and other sources of information can help you develop your critical thinking skills. By exposing yourself to different perspectives and ideas, you can learn to analyze and evaluate information more effectively.

2. Practice mindfulness: Mindfulness is the practice of being present in the moment and paying attention to your thoughts and feelings. By being more aware of your thoughts and emotions, you can better understand and evaluate them. This can help you develop a more objective and rational approach to thinking.

3. Engage in open-minded discussions: Join a discussion group or debate club where you can engage in open-minded discussions with others. This can help you develop your critical thinking skills by challenging your assumptions and encouraging you to consider different perspectives.
\\
{\bf LLaMA-Adapter:}
1. Read widely and critically.
2. Engage in discussions with people who have different perspectives.
3. Ask questions and seek out answers.
\end{minipage}
\end{framed}
\vspace{-5mm}

\begin{framed}
\centering
 \begin{minipage}{\textwidth}
{\bf Instruction:} As a superhero, how would you explain your origin story to a curious child? \\
{\bf LLaMA-Adapter V2:} As a superhero, I have a very interesting and exciting origin story to share with you. Once upon a time, there was a young boy named John who lived in a small town. John was always curious and wanted to help others, so he spent his days exploring the world around him and learning about different cultures and people. One day, while he was out exploring, he stumbled upon a mysterious cave. Inside the cave, John found a strange device that had the power to give him incredible abilities. With these abilities, John could fly, shoot lasers from his eyes, and even turn invisible.

John knew that he had to use these powers for good, so he decided to become a superhero and use his abilities to protect his town and the people who lived there. He started to train every day, learning how to use his powers and how to fight against evil. Eventually, John became the superhero known as "The Invisible Man" and he used his powers to help others and make the world a better place.

That's my origin story, and I'm sure you'll love it!
\\
{\bf LLaMA-Adapter:} Once upon a time, there was a young boy who had a special power. He used his power to help people in need and protect the innocent. He became known as the superhero, and he used his powers to fight evil and protect the world.
\end{minipage}
\end{framed}
\vspace{-5mm}

\begin{framed}
\centering
 \begin{minipage}{\textwidth}
{\bf Instruction:} Write a regex to remove all the HTML tags in a python string. \\
{\bf LLaMA-Adapter V2:}
Here is a simple regex that can be used to remove all HTML tags from a string in Python:

\begin{lstlisting}
import re

string = "This is a <strong>test</strong> string"

# Remove all HTML tags from the string
result = re.sub(r'<.*?>', '', string)

# Print the result
print(result)
\end{lstlisting}

This will output:

\begin{lstlisting}
This is a test string
\end{lstlisting}
{\bf LLaMA-Adapter:} 
\begin{lstlisting}
re.sub(r'<[^>]*>', '', str)
\end{lstlisting}
\end{minipage}
\end{framed}

\end{tabular}%
\vspace{-5mm}
\caption{\textbf{Comparisons of Instruction-following Ability} between LLaMA-Adapter V2 and LLaMA-Adapter.}
\label{tab:language_instruction_comparison}
\end{table*}

\vspace{0.2cm}
\subsection{Early Fusion of Visual Knowledge}
\label{subsec:early_fusion}
To avoid interference between visual and language fine-tuning, we propose a simple early fusion strategy to prevent direct interactions between the input visual prompts and adaptation prompts.
In LLaMA-Adapter, the input visual prompts are sequentially encoded by a frozen visual encoder with a learnable visual projection layer, and then added to the adaptation prompts at every inserted layer.
In LLaMA-Adapter V2, we instead inject the encoded visual tokens and adaptation prompts to different Transformer layers without fusing them together, which is shown in Fig.~\ref{fig:early_fusion}. For the dataset-shared adaptation prompts, we follow LLaMA-Adapter by inserting them at the last $L$ layers (\emph{e.g.}, $L$=30). For the input visual prompts, we directly concatenate them with the word tokens at the first Transformer layer with the zero-initialized attention, other than adding them to the adaptation prompts. Together with the proposed joint training, this simple early fusion strategy of visual tokens can effectively resolve the conflict between the two types of fine-tuning targets. This contributes to a parameter-efficient LLaMA-Adapter V2 with superior multi-modal reasoning capabilities.

\vspace{0.2cm}
\subsection{Integration with Experts}
\label{subsec:with_experts}



Recent visual instruction models such as MiniGPT-4~\cite{zou2022xdecoder} and LLaVA~\cite{liu2023visual} require massive-scale image-text training to connect visual models and LLMs. In contrast, our LLaMA-Adapter V2 fine-tunes on much smaller-scale common image captioning data~\cite{chen2015microsoft}, making it more data-efficient. However, the image understanding ability of our approach is relatively weak, leading to occasional inaccurate or unrelated responses.
Rather than collecting more image-text data or adopting stronger multi-modal modules, we propose integrating expert systems, such as captioning, OCR, and search engines, to supplement LLaMA-Adapter V2 with additional visual reasoning proficiency.



As shown in Fig.~\ref{fig2}, we utilize expert systems, such as captioning, detection, and OCR, to enhance the visual instruction-following capabilities of LLaMA-Adapter V2.
Given an input image, we encode its visual context using a pre-trained visual encoder and ask an expert system to produce a caption as the textual context. 
In our default implementation, we adopt LLaMA-Adapter pre-trained on COCO Caption~\cite{chen2015microsoft} as the expert system, since it can generate short and accurate image descriptions. 
However, it is worth noting that any image-to-text model or even a search engine can serve as the expert system here. 
Our approach allows us to easily switch among different expert systems based on the specific downstream task at hand.


\section{Experiments}
\label{sec:exps}

\begin{figure}[ht]
    \centering
    \subfloat[The total quality scores compared with ChatGPT.]{\includegraphics[width=\textwidth]{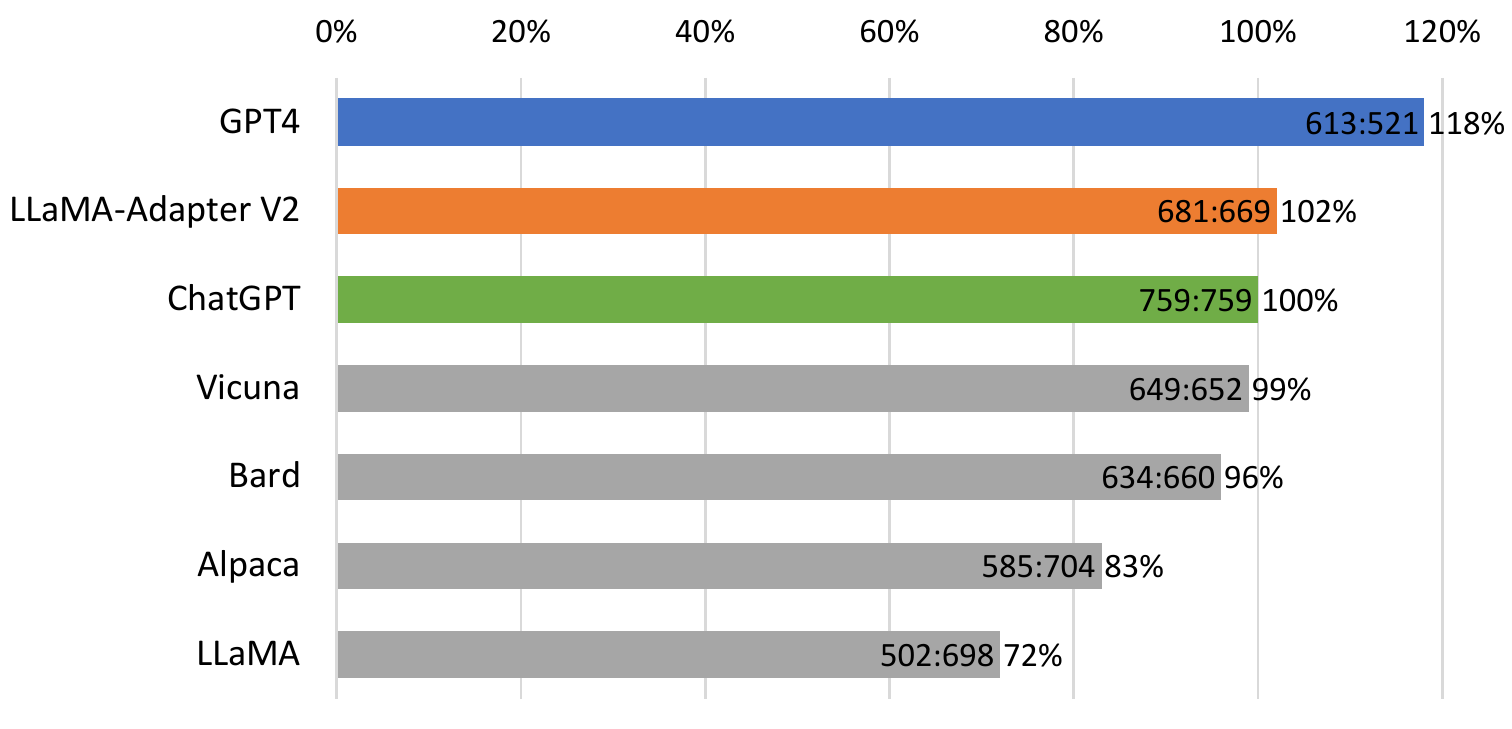}} \\
    \subfloat[Left: Vicuna \emph{vs.} ChatGPT. Right: LLaMA-Adapter V2 \emph{vs.} ChatGPT.]{\includegraphics[width=0.9\textwidth]{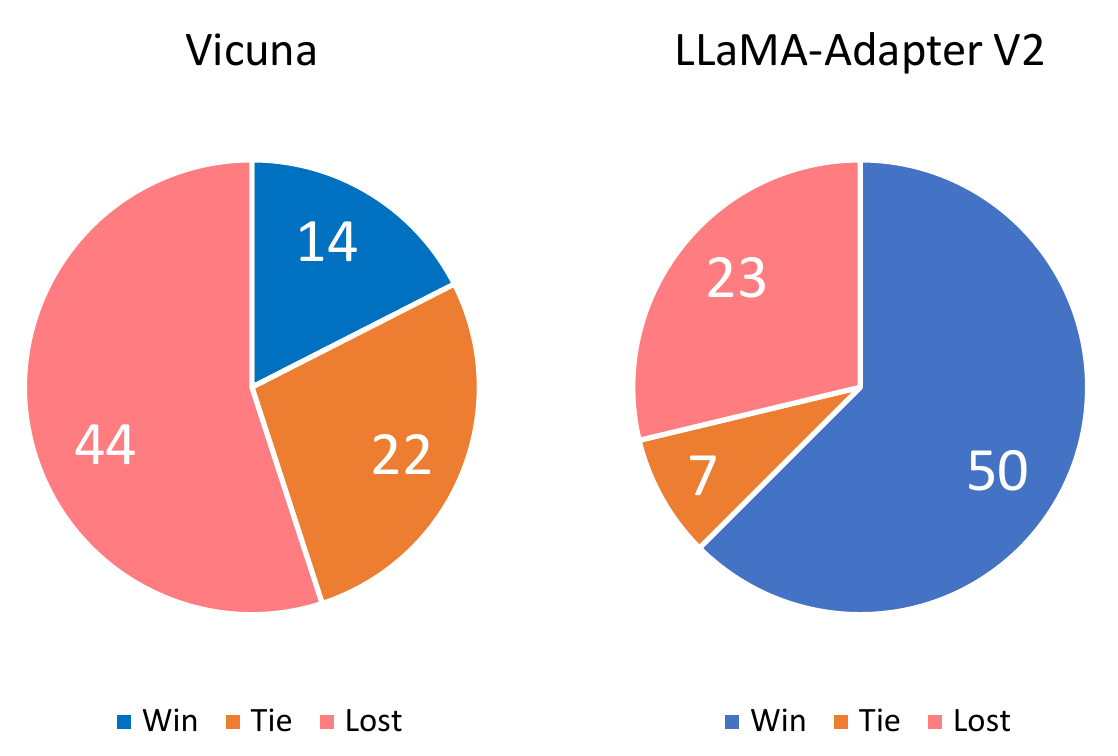}}
    \caption{\textbf{Response Quality Comparisons} assessed by GPT4. The comparison baseline is ChatGPT. We use GPT4 to assess the response quality of different methods on 80 questions~\cite{vicuna2023}. Note that our model is based on LLaMA-65B, while Vicuna is built on LLaMA-13B. But we only fine-tune \textbf{14M} parameters, while Vicuna requires fine-tuning the whole 13B parameters.}
    \label{fig:gpts_scores_all}
\end{figure}

\begin{figure*}[tbhp]
    \centering
    \resizebox{\textwidth}{!}{%
\begin{tabular}{ll}
\textbf{\textit{\textcolor[HTML]{548235}{Success Case 1}}}\\
\multirow{2}{*}{
\includegraphics[width=0.12\columnwidth]{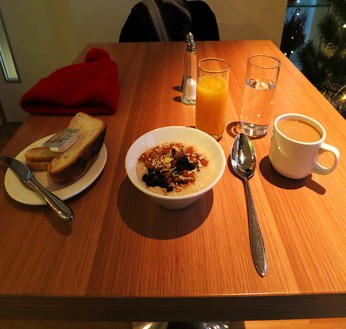}
} & \begin{minipage}[t]{\columnwidth}
\textbf{V1:} A bowl of cereal, a cup of coffee and a sandwich on a table.\end{minipage}\\
& \begin{minipage}[t]{\columnwidth}\textbf{V2:} The image features a dining table set up with a variety of food items, including a bowl of fruit, a cup of coffee, and a plate of pastries. The table is set with a knife, fork, and a spoon, suggesting that the food is ready to be consumed. The dining table is situated in a kitchen, with a chair placed near the table. The chair is positioned next to a small table, which may be used for additional food items or drinks. The room is well-lit, with a lamp illuminating the area.\end{minipage} \\
\textbf{\textit{\textcolor[HTML]{548235}{Success Case 2}}}\\
\multirow{2}{*}{
\includegraphics[width=0.12\columnwidth]{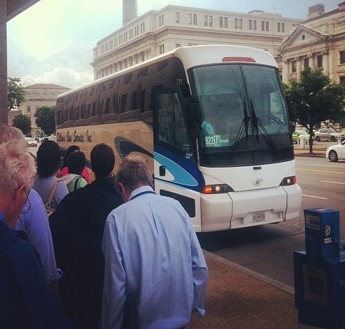}
} & \begin{minipage}[t]{\columnwidth}
\textbf{V1:} A group of people waiting to board a bus.\end{minipage}\\
& \begin{minipage}[t]{\columnwidth}\textbf{V2:} In the image, a large bus is parked on the side of the road, with several people standing nearby. There are at least three people standing near the bus, with one person holding a backpack and another person holding a bag. A few more people can be seen in the background, walking towards the bus. The bus is painted white, and its front end is visible, with a few people standing near the front. A few cars are also visible in the background, adding to the busy street scene.\end{minipage} \\
\textbf{\textit{\textcolor[HTML]{C00000}{Failure Case}}}\\
\multirow{2}{*}{
\includegraphics[width=0.12\columnwidth]{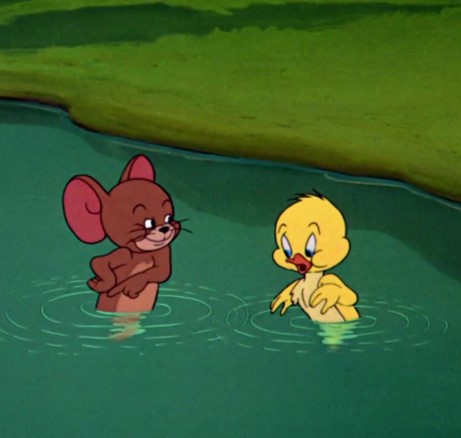}
} & \begin{minipage}[t]{\columnwidth}
\textbf{V1:} A mouse and a duck in a body of water.\end{minipage}\\
& \begin{minipage}[t]{\columnwidth}\textbf{V2:} In the image, two men are standing on a dock near a lake or river. One man is wearing a blue shirt and the other is wearing a white shirt. Both men are looking down and appear to be fishing. They are standing near a small rocky area and are in the water up to their knees.   In the image, there is also a dog, a yellow labrador retriever, swimming in the water near the men. The dog appears to be swimming towards the men, perhaps attracted by the presence of the fishermen. The dog is wearing a red collar and appears to be enjoying its time in the water with the two men.\end{minipage} \\
\end{tabular}%
}
\caption{\textbf{Comparisons of Image Captioning Results} between LLaMA-Adapter and LLaMA-Adapter V2.\vspace{0.1cm}}
\label{tab:caption_demo}
\end{figure*}

\subsection{Experimental Setups}
\paragraph{Training Data.} As shown in Tab.~\ref{tab:param_compare_with_minigpt4_llava}, we train LLaMA-Adapter V2 on 52K single-turn instruction data from GPT-4-LLM~\cite{peng2023instruction} and 567K captioning data from COCO Caption~\cite{chen2015microsoft}. We do not use any visual instruction data like MiniGPT-4~\cite{zou2022xdecoder} and LLaVA~\cite{liu2023visual}. Besides, we also train a chatbot system using 80K conversation data collected by ShareGPT~\cite{sharegpt}.

\paragraph{Implementation Details.}
For the LLaMA-7B model with 32 Transformer layers, we insert static adaptation prompts into the last 31 layers. 
Additionally, we append dynamic visual prompts to the first layer, with a prompt length set to 20. 
All the parameters in normalization layers, linear layer bias and scale are set to be updated during training, while the remaining parameters in LLaMA are kept frozen.

\subsection{Stronger Language Instruction Model}

With the proposed bias tuning strategy and high-quality instruction data~\cite{peng2023instruction}, LLaMA-Adapter V2 was able to further enhance the instruction-following capability of LLaMA. 
As shown in Tab.~\ref{tab:language_instruction_comparison}, LLaMA-Adapter V2 can provide comprehensive answers to human instructions
as well as the detailed explanations of the answers, while LLaMA-Adapter only delivers relatively short answers.

Given that bias tuning involves more learnable parameters for knowledge updating, it is possible to build a chatbot system that requires a deeper understanding of language context. 
By training LLaMA-Adapter V2 on 80K conversation data~\cite{sharegpt}, we developed a stronger chatbot model. 
Fig.~\ref{tab:chat_demo_7b} presents a chatbot example using a 7 billion model where the system is able to answer our questions, but its understanding of context is not very accurate. 
By scaling the model to 65 billion (Fig.~\ref{tab:chat_demo_65b}), the chatbot becomes more powerful and answers our questions very well. In Fig~\ref{fig:gpts_scores_all}, we also use GPT4 to assess the response quality, where LLaMA-Adapter V2 even beats ChatGPT on the total scores and 50/80 questions.

\begin{table}[t]
\centering
\small
\begin{tabular}{c|cc|cc}
\toprule
\multirow{2}{*}{Model} & \multicolumn{2}{c|}{Data Scale} & \multicolumn{2}{c}{COCO Caption} \\
                       & PT      & FT      & BLEU@4          & CIDEr          \\
\midrule
ClipCap~\cite{mokady2021clipcap}                & 0             & 0.6M           & 33.5            & 113.1          \\
BLIP~\cite{li2022blip}                   & 14M           & 0.6M           & 40.4            & 136.7          \\
BLIP2~\cite{li2023blip}                  & 129M          & 0.6M           & 43.7            & 145.3          \\
\midrule
LLaMA-Adapter V2         & 0             & 0.6M           & 36.2            & 122.2\\
\bottomrule
\end{tabular}%
\caption{\textbf{Comparisons on COCO Caption.} BLIP and BLIP2 require pretraining (PT) on COCO Caption~\cite{chen2015microsoft}, Visual Genome~\cite{krishna2017visual}, Conceptual Captions~\cite{sharma2018conceptual,changpinyo2021conceptual} and LAION~\cite{schuhmann2021laion}, while ClipCap and our approach only fine-tune (FT) the model on COCO Caption.}
\label{tab:coco_cap}
\end{table}

\begin{figure*}[th]
\centering
\resizebox{\textwidth}{!}{%
\begin{tabular}{ll}
\multirow{2}{*}{
\includegraphics[width=0.17\columnwidth]{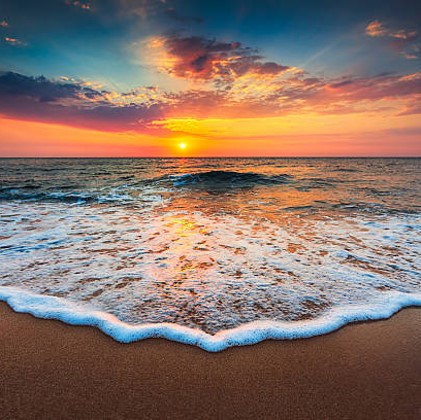}
} & \begin{minipage}[t]{\columnwidth}
\textbf{Instruction:} Explain why this image is beautiful.\end{minipage}\\
& \begin{minipage}[t]{\columnwidth}\textbf{Answer:} This image is beautiful because it captures the serene and peaceful atmosphere of the beach, with the sun setting in the background. The image features a beautiful, golden sunset over the ocean, creating a stunning backdrop for the solitary figure walking along the sandy beach. The combination of the warm, golden light of the sunset, the blue water of the ocean, and the white sandy beach creates a breathtaking and calming scene. The image also highlights the vastness of the ocean, with the waves crashing against the shore, creating a soothing sound that adds to the tranquility of the setting. The image conveys a sense of relaxation and tranquility, making it a perfect representation of a beautiful beach scene.\end{minipage} \\
& \\
\multirow{2}{*}{
\includegraphics[width=0.17\columnwidth]{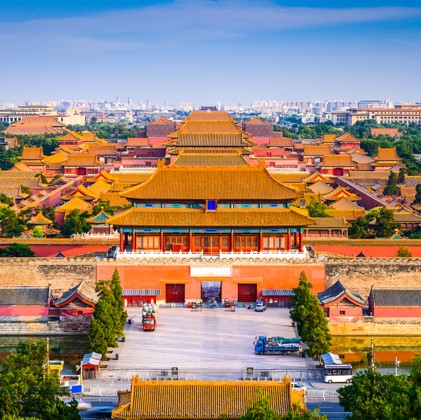}
} & \begin{minipage}[t]{\columnwidth}
\textbf{Instruction:} What city might this picture be taken from? Can you give me some suggestions if I want to visit this city?\end{minipage}\\
& \begin{minipage}[t]{\columnwidth}\textbf{Answer:} The image might be taken from Beijing, China. Beijing is the capital city of China and one of the most populous cities in the world. It is known for its rich history and culture, as well as its modern infrastructure and architectural landmarks. Some popular attractions in Beijing include the Forbidden City, Temple of Heaven, and the Great Wall of China, among others. If you want to visit Beijing, it is recommended to explore its historical and cultural heritage, as well as its modern developments. \end{minipage} \\ \\
& \\
\multirow{2}{*}{
\includegraphics[width=0.17\columnwidth]{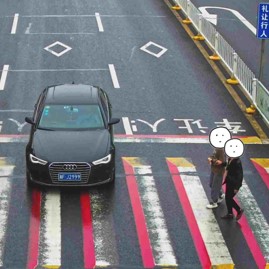}
} & \begin{minipage}[t]{\columnwidth}
\textbf{Instruction:} What should the car do here?\end{minipage}\\
& \begin{minipage}[t]{\columnwidth}\textbf{Answer:} The car should follow the pedestrian crossing the street, as indicated by the red light. The car should stop at the red light and wait for the pedestrian to cross the street safely before proceeding. This is a common practice to ensure the safety of pedestrians and drivers alike. \end{minipage} \\ \\ \\ \\

\end{tabular}%
}
\caption{\textbf{The Visual Understanding Examples} of LLaMA-Adapter V2.}
\label{tab:visual_understanding}
\end{figure*}

\subsection{Visual Instruction Model}

LLaMA-Adapter is primarily a language instruction model or a close-set vision-language model, while LLaMA-Adapter V2 is a powerful visual instruction model that is jointly trained on caption and language-only instruction data.
In this section, we study LLaMA-Adapter V2's image captioning ability and demonstrate how it can be extended to general-purpose multi-modal understanding systems like GPT-4. In addition, we incorporate expert systems to further improve LLaMA-Adapter V2's visual instruction-following capability.

\paragraph{Image Captioning.}
As discussed in Sec.~\ref{sec:preliminary}, LLaMA-Adapter supports multi-modal inputs by simply adding visual features to adaptation prompts. After fine-tuning on COCO Caption dataset, we transform LLaMA-Adapter into a strong image captioner. 
In Tab.~\ref{tab:coco_cap}, we can observe that LLaMA-Adapter achieves comparable results to BLIP~\cite{li2022blip,li2023blip}, even without pre-training on large-scale image-text data. However, LLaMA-Adapter cannot reuse the language modeling ability of the LLM since it is sensitive to specific prompts, such as ``Generate caption for this image".

By employing early fusion and joint training, LLaMA-Adapter V2 has become a powerful visual instruction model that can simultaneously perform language instruction-following and image captioning. 
We provide some examples in Fig.~\ref{tab:caption_demo} to compare the image captioning results between LLaMA-Adapter and LLaMA-Adapter V2. 
For an given image, LLaMA-Adapter can only produce a short, close-form image description, while LLaMA-Adapter V2 is able to generate natural and detailed image descriptions.

In addition, we note that LLaMA-Adapter V2 may not always generate accurate image description. 
As seen in the failure case presented in Fig.~\ref{tab:caption_demo}, we intentionally chose an out-of-distribution example (cartoon picture) for testing. LLaMA-Adapter V2 was unable to comprehend the image and generated incorrect descriptions.
The possible reason for this failure is the lack of an image-text alignment stage, which models like MiniGPT-4 and LLaVA have incorporated. 
This motivates us to employ additional expert systems to enhance the image understanding ability.

\paragraph{Visual Understanding.}
Thanks to our proposed joint training and early fusion techniques, LLaMA-Adapter V2 exhibits exceptional multi-modal understanding capabilities. 
As shown in Fig.~\ref{tab:visual_understanding}, when presented with a prompt about the image content, such as in the form of ``why is ...'' and ``what should ...'', our model is able to seamlessly integrate the visual information with the language context, allowing for more sophisticated reasoning and decision-making. From the examples, we can see that our model is able to identify and explain the specific object or feature in the image that the question is referring to, and provide relevant information or suggestions based on the context. Moreover, our model is able to extrapolate and provide a plausible guess or explanation even when the question goes beyond the direct information presented in the image. 
These results demonstrate the effectiveness of LLaMA-Adapter V2 in addressing the challenge of interference between image-text pairs and instruction data, and highlight the potential of LLaMA-Adapter V2 in various real-world applications that require both language and visual understanding.


\begin{figure*}[thbp]
\centering
\resizebox{\textwidth}{!}{%
\begin{tabular}{ll}
\multirow{3}{*}{
\includegraphics[width=0.17\columnwidth]{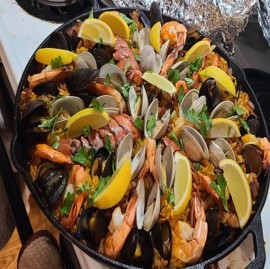}
} & \begin{minipage}[t]{\columnwidth}
\textbf{Instruction:} Could you list the steps for cooking the dish shown in the image?\end{minipage}\\
& \begin{minipage}[t]{\columnwidth}\textbf{w/o Context:} Yes, the steps for cooking the dish shown in the image can be summarized as follows:  1. Boil the octopus in salted water to tenderize the meat.  2. Remove the octopus from the boiling water and drain the excess liquid.  3. Allow the octopus to cool down to room temperature.  4. Cut the octopus into small pieces, preferably 1-2 inches in size.  5. In a large skillet, heat the olive oil and cook the octopus pieces until they turn golden brown.  6. Add the garlic and lemon juice to the skillet and continue to cook the octopus for a few more minutes.  7. Season the octopus with salt and pepper, to taste.  8. Serve the cooked octopus on a plate with the lemon wedges and garnish with the chopped parsley.\end{minipage} \\
& \begin{minipage}[t]{\columnwidth}
\textbf{w/ Context:} [\textit{Caption expert: A seafood dish with shrimp, octopus, and lobster.}] The steps for cooking the seafood dish with shrimp, octopus, and lobster are as follows:
1. Start by preparing the seafood.
2. Heat up the oil in a pan or skillet.
3. Add the shrimp, octopus, and lobster to the pan.
4. Cook the seafood over medium heat for 10 minutes.
5. Season the dish with salt and pepper to taste.
6. Serve the dish with a side of rice, vegetables, or other starch.
7. Enjoy! \end{minipage}\\
\\
\multirow{3}{*}{
\includegraphics[width=0.17\columnwidth]{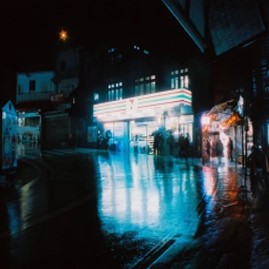}
} & \begin{minipage}[t]{\columnwidth}
\textbf{Instruction:} How to shoot such a picture using a film camera at night? Do you have some suggestions?\end{minipage}\\
& \begin{minipage}[t]{\columnwidth}\textbf{w/o Context:} To capture the image of the neon sign in the rain, a film camera would need to be set up with a long exposure time to allow the light from the sign to be captured. This would require the photographer to use a tripod to keep the camera steady and to manually adjust the shutter speed to the desired exposure time. The photographer would also need to use a slow shutter speed to capture the light from the sign, which could range from 1 second to several minutes, depending on the desired effect.  Additionally, the photographer would need to use a wide aperture setting to ensure that the entire sign is in focus, as well as a low ISO setting to minimize noise and preserve the image quality.\end{minipage} \\
& \begin{minipage}[t]{\columnwidth}
\textbf{w/ Context:} [\textit{Caption expert: A wet street with a building and a car.}] To capture a nighttime scene with a wet street, a film camera, and a car, the photographer would need to use a slow shutter speed to allow more light to enter the camera's sensor. This would result in a blurred image of the moving car, but the building and the street would be sharply focused. The photographer could also use a tripod to steady the camera and reduce the shake caused by handheld shooting. Additionally, the photographer could use a flash to illuminate the car and the street, creating a more dramatic effect. \end{minipage}\\
\end{tabular}%
}
\caption{\textbf{Visual Understanding with the help of Caption Experts.} The caption expert is LLaMA-Adapter V1 trained on COCO Caption~\cite{chen2015microsoft}.}
\label{tab:integration_w_caption}
\end{figure*}
\begin{figure*}[t]
\centering
\small
\resizebox{\textwidth}{!}{%
\begin{tabular}{ll}
\multirow{2}{*}{
\includegraphics[width=0.35\columnwidth]{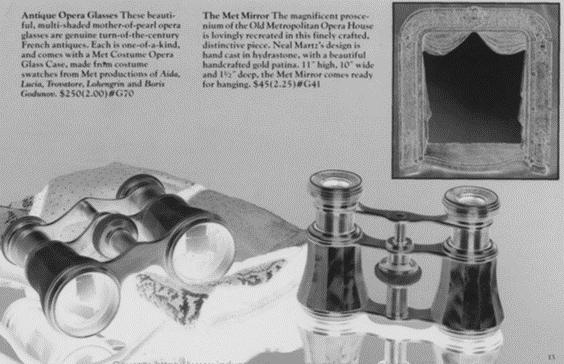}
} & \begin{minipage}[t]{0.62\columnwidth}
\textbf{Question:} What is the cost of the Antique Opera glasses? Do you think it's worth buying the Antique Opera glasses and can you explain the reason in detail?\end{minipage}\\
& \begin{minipage}[t]{0.62\columnwidth}\textbf{OCR  Context:} 1. Antique Opera Glasses These beauti-  2. ful, multi-shaded mother-of-pearl opera 3. The Met Mirror The magnificent prosce- 4. glasses are genuine turn-of-the-century 5. nium of the Old Metropolitan Opera House 6. French antiques. Each is one-of-a-kind, 7. is lovingly recreated in this finely crafted, 8. and comes with a Met Costume Opera 9. distinctive piece. Neal Martz's design is 10. hand cast in hydrastone, with a beautiful 11. Glass Case, made from costume 12. swatches from Met productions of Aida, 13. handcrafted gold patina. 11 high, 10 wide 14. Lucia, Trovatore, Lohengrin and Boris 15. and 11/2 deep, the Met Mirror comes ready 16. Godunov. \$250(2.00) \#G70 17. for hanging. \$45(2.25) \#G41 18. 13\end{minipage} \\

\multicolumn{2}{l}{
 \begin{minipage}[t]{\columnwidth}
\textbf{Language Instruction Model:} I'm sorry, but I am an AI language model and I don't have the ability to access images or prices. Could you please provide me with the price of the Antique Opera glasses so that I can provide you with an accurate response? \end{minipage}
}
\\

\multicolumn{2}{l}{
\begin{minipage}[t]{\columnwidth}
\textbf{Visual Instruction Model w/o Context:} The cost of the Antique Opera glasses is not mentioned in the image. However, the image suggests that the glasses are being sold in a store, which means that the cost may vary depending on the location, the vendor, and the specific glasses being sold.  It is important to note that the value of the Antique Opera glasses is subjective and may depend on the individual's taste, preference, and budget. Some people may find the antique design and the history behind the glasses appealing and may be willing to pay a higher price for them.\end{minipage}
}
\\

\multicolumn{2}{l}{
\begin{minipage}[t]{\columnwidth}
\textbf{Visual Instruction Model w/ Context:} The cost of the Antique Opera glasses is \$250 (2.00) and it is worth buying because they are genuine turn-of-the-century French antiques that are lovingly recreated in this finely crafted, hand cast in hydrastone, with a beautiful gold patina. The Met Mirror comes ready for hanging and is \$45 (2.25).\end{minipage}
}

\end{tabular}%
}
\caption{\textbf{Visual Understanding with the help of OCR Experts.} The example and OCR context are from DocVQA~\cite{mathew2021docvqa}.\vspace{0.2cm}}
\label{tab:integration_w_ocr}
\end{figure*}

\paragraph{Integration with Experts.}



To further enhance LLaMA-Adapter V2's visual understanding capabilities, we integrate visual expert models during inference to provide additional textual contexts based on the input image. 
The examples of using LLaMA-Adapter~\cite{llamaadapter2023} as the caption expert are shown in Fig.~\ref{tab:integration_w_caption}. 
By including a short caption of the image, our LLaMA-Adapter V2 produces more precise and detailed responses about the visual contents in the image.
Another example using the OCR expert from DocVQA~\cite{mathew2021docvqa} is presented in Fig.~\ref{tab:integration_w_ocr}. 
By utilizing the detected texts in the image, our model receives more specific cues, such as the price of the glasses, and produces accurate answers to the question. 
The integration of experts thus largely extends the power of LLaMA-Adapter V2 and fully unleashes its multi-modal reasoning potential.

\section{Conclusion}
\label{sec:conclusion}
In this work, we propose LLaMA-Adapter V2, a parameter-efficient visual instruction tuning system.
Through joint training on image-text pairs and instruction-following data, we observe that LLaMA-Adapter V2 can convert pre-trained LLMs into a zero-shot visual instruction model. 
With the assistance of simple bias-tuning strategy and early-fusion of visual features, LLaMA-Adapter V2's zero-shot visual instruction-following ability is further improved by alleviating the interference between image-text pairs and instruction-following data. 
At the same time, LLaMA-Adapter V2 achieves stronger language instruction-following performance and even possesses multi-turn dialog ability like a chatbot.
To address the inaccurate image description problem of LLaMA-Adapter V2, we integrate expert visual systems, such as OCR and image captioner, to provide more accurate image-related information. 
While LLaMA-Adapter V2 can incorporate expert visual systems and perform visual instruction-following in a zero-shot manner, it still lags behind LLaVA in terms of visual understanding capability and is susceptible to inaccurate information provided by the expert systems.
In the future, we plan to explore the integration of more expert systems, and fine-tune LLaMA-Adapter V2 with a multi-modal instruction dataset or other PEFT methods (\emph{e.g.}, LoRA) to further enhance its visual instruction-following capabilities.

\begin{figure*}[ht]
    \centering
    \includegraphics[width=\textwidth]{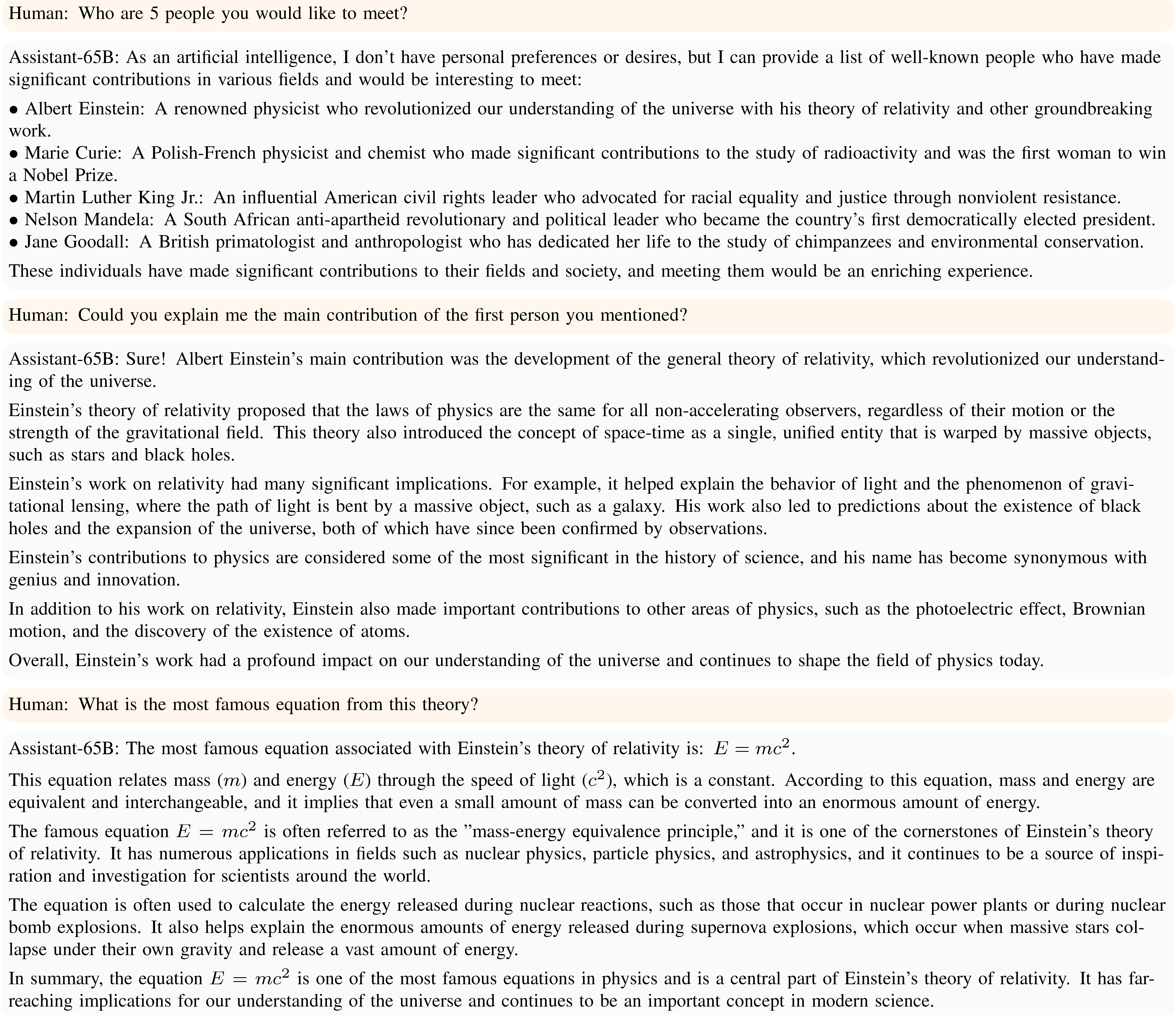}
    \caption{\textbf{A Chatting Example using 65B LLaMA-Adapter V2.}}
    \label{tab:chat_demo_65b}
\end{figure*}

\begin{figure*}[ht]
    \centering
    \includegraphics[width=\textwidth]{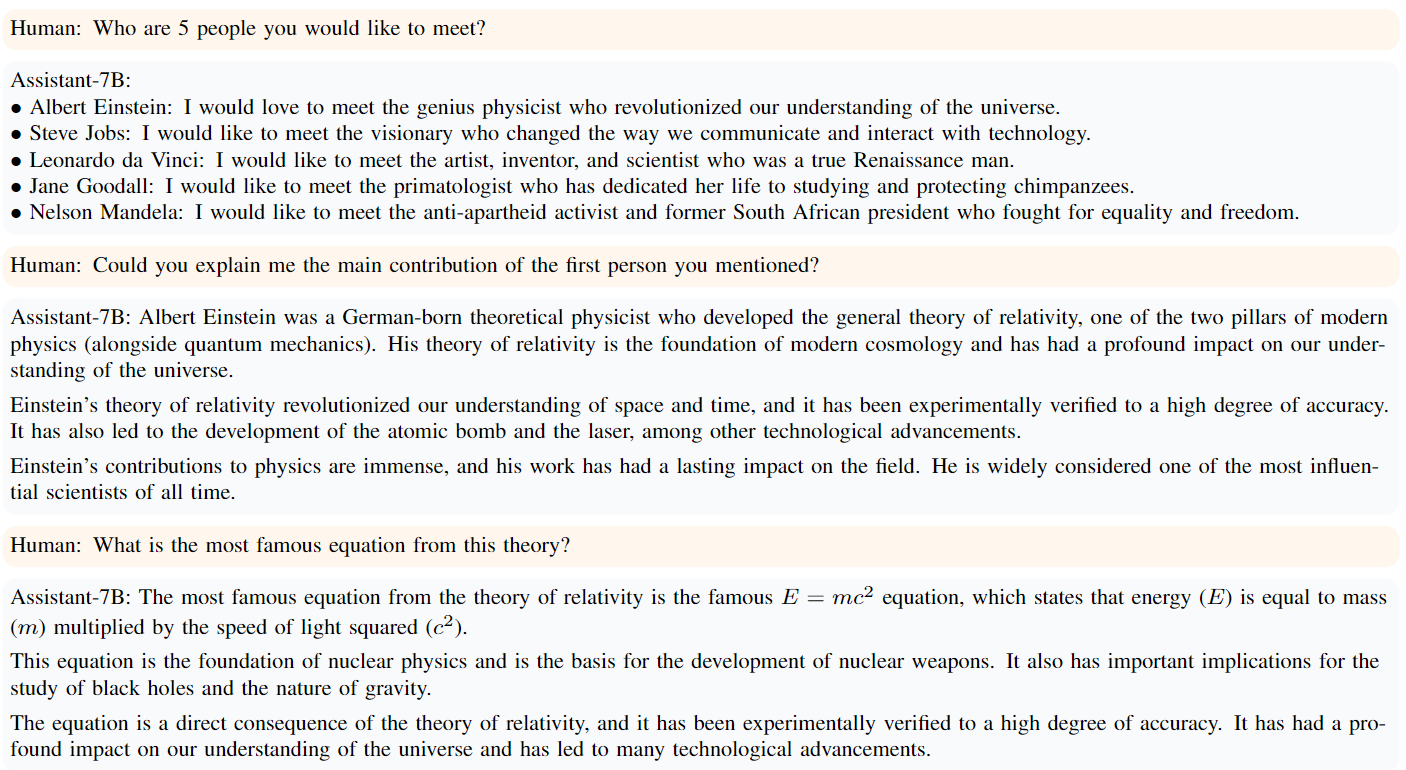}
    \caption{\textbf{A Chatting Example using 7B LLaMA-Adapter V2.}}
    \label{tab:chat_demo_7b}
\end{figure*}

{\small
\bibliographystyle{ieee_fullname}
\bibliography{egbib}
}

\end{document}